\documentclass[10pt,twocolumn,letterpaper]{article}

\usepackage{cvpr}
\usepackage{times}
\usepackage{epsfig}
\usepackage{graphicx}
\usepackage{amsmath}
\usepackage{amssymb}
\usepackage{subcaption}

\usepackage{url}
\usepackage{relsize}
\usepackage{xcolor,colortbl}
\usepackage{multirow}
\usepackage{wrapfig}
\usepackage{bbm}
\usepackage[labelfont=bf]{caption}
\usepackage{tabularx}

\usepackage[pagebackref=true,breaklinks=true,letterpaper=true,colorlinks,bookmarks=false]{hyperref}

\cvprfinalcopy 

\def\cvprPaperID{3485} 

\newcommand{\rxs}{69.0}
\newcommand{\rsmall}{73.7}
\newcommand{\rmed}{76.8}
\newcommand{\rlarge}{77.8}
\newcommand{\rxl}{78.6}

\newcommand{\sxs}{91}
\newcommand{\ssmall}{81}
\newcommand{\smed}{67}
\newcommand{\slarge}{59}
\newcommand{\sxl}{40}

\newcommand{\inet}{19.7}
\newcommand{\inone}{16.0}

\begin{document}

\title{\vspace{-2cm}YOLO9000: \\
Better, Faster, Stronger\vspace{-.25cm}}

\author{Joseph Redmon$^{* \dag}$, Ali Farhadi$^{* \dag}$\\
\small{University of Washington$^*$, Allen Institute for AI$^\dag$}\\ \url{http://pjreddie.com/yolo9000/}}

\maketitle
\begin{abstract}
\vspace{-.25cm}
We introduce YOLO9000, a state-of-the-art, real-time object detection system that can detect over 9000 object categories. First we propose various improvements to the YOLO detection method, both novel and drawn from prior work. The improved model, YOLOv2, is state-of-the-art on standard detection tasks like \textsc{Pascal} VOC and COCO. Using a novel, multi-scale training method the same YOLOv2 model can run at varying sizes, offering an easy tradeoff between speed and accuracy. At \smed{} FPS, YOLOv2 gets \rmed{} mAP on VOC 2007. At \sxl{} FPS, YOLOv2 gets \rxl{} mAP, outperforming state-of-the-art methods like Faster R-CNN with ResNet and SSD while still running significantly faster. Finally we propose a method to jointly train on object detection and classification. Using this method we train YOLO9000 simultaneously on the COCO detection dataset and the ImageNet classification dataset. Our joint training allows YOLO9000 to predict detections for object classes that don't have labelled detection data. We validate our approach on the ImageNet detection task. YOLO9000 gets \inet{} mAP on the ImageNet detection validation set despite only having detection data for 44 of the 200 classes. On the 156 classes not in COCO, YOLO9000 gets \inone{} mAP. But YOLO can detect more than just 200 classes; it predicts detections for more than 9000 different object categories. And it still runs in real-time.
\end{abstract}

\begin{figure}[]
      \centering
        \includegraphics[width=\linewidth]{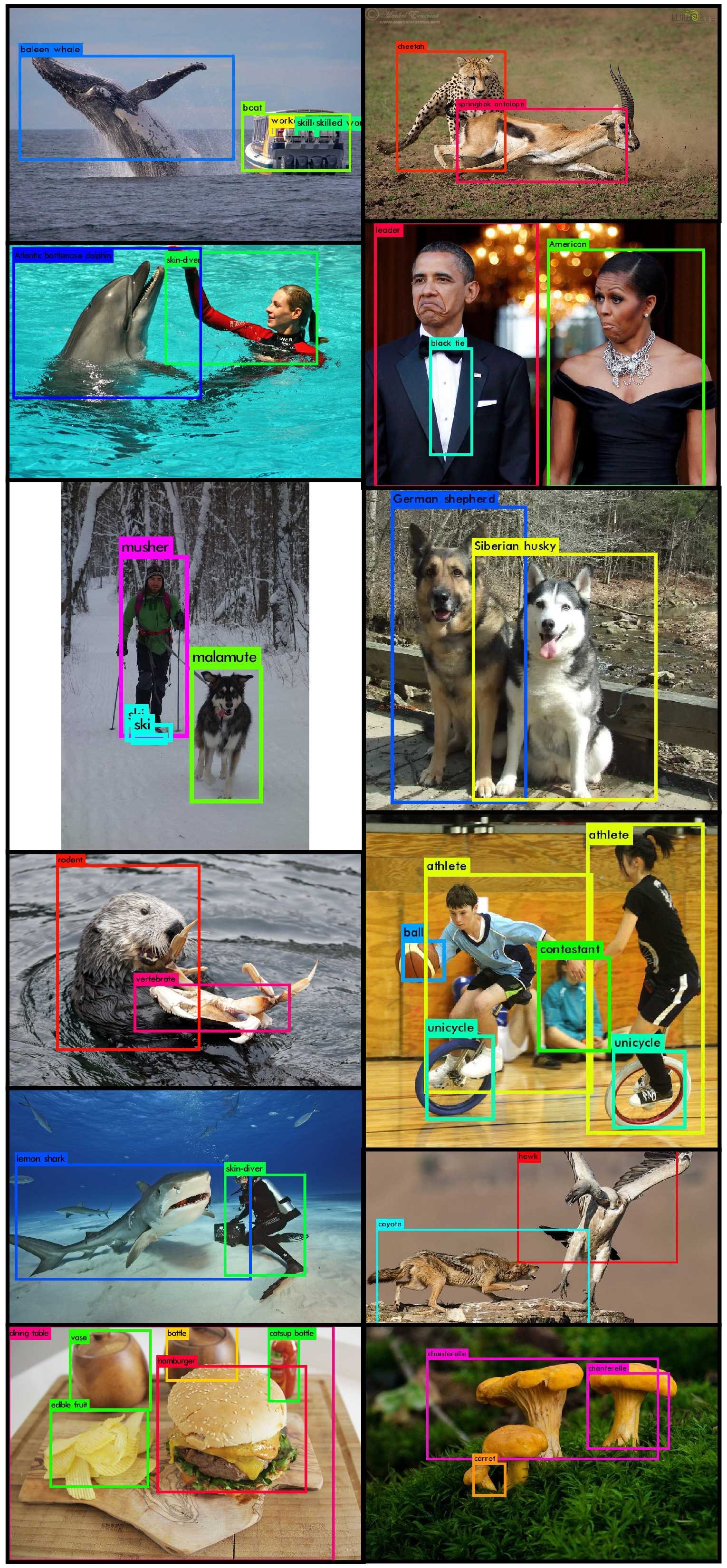}
      \caption{\small \textbf{YOLO9000.} YOLO9000 can detect a wide variety of object classes in real-time.}
      \label{y9k}
   \end{figure}

\section{Introduction}

General purpose object detection should be fast, accurate, and able to recognize a wide variety of objects. Since the introduction of neural networks, detection frameworks have become increasingly fast and accurate. However, most detection methods are still constrained to a small set of objects.

Current object detection datasets are limited compared to datasets for other tasks like classification and tagging. The most common detection datasets contain thousands to hundreds of thousands of images with dozens to hundreds of tags \cite{voc} \cite{coco} \cite{deng2009imagenet}. Classification datasets have millions of images with tens or hundreds of thousands of categories \cite{tag} \cite{deng2009imagenet}.

We would like detection to scale to level of object classification. However, labelling images for detection is far more expensive than labelling for classification or tagging (tags are often user-supplied for free). Thus we are unlikely to see detection datasets on the same scale as classification datasets in the near future.

We propose a new method to harness the large amount of classification data we already have and use it to expand the scope of current detection systems. Our method uses a hierarchical view of object classification that allows us to combine distinct datasets together.

We also propose a joint training algorithm that allows us to train object detectors on both detection and classification data. Our method leverages labeled detection images to learn to precisely localize objects while it uses classification images to increase its vocabulary and robustness.

Using this method we train YOLO9000, a real-time object detector that can detect over 9000 different object categories. First we improve upon the base YOLO detection system to produce YOLOv2, a state-of-the-art, real-time detector. Then we use our dataset combination method and joint training algorithm to train a model on more than 9000 classes from ImageNet as well as detection data from COCO.

All of our code and pre-trained models are available online at \url{http://pjreddie.com/yolo9000/}.

\section{Better}

YOLO suffers from a variety of shortcomings relative to state-of-the-art detection systems. Error analysis of YOLO compared to Fast R-CNN shows that YOLO makes a significant number of localization errors. Furthermore, YOLO has relatively low recall compared to region proposal-based methods. Thus we focus mainly on improving recall and localization while maintaining classification accuracy.

Computer vision generally trends towards larger, deeper networks \cite{resnet} \cite{inception} \cite{vgg}. Better performance often hinges on training larger networks or ensembling multiple models together. However, with YOLOv2 we want a more accurate detector that is still fast. Instead of scaling up our network, we simplify the network and then make the representation easier to learn. We pool a variety of ideas from past work with our own novel concepts to improve YOLO's performance. A summary of results can be found in Table \ref{analysis}.

\textbf{Batch Normalization.} Batch normalization leads to significant improvements in convergence while eliminating the need for other forms of regularization \cite{batch}. By adding batch normalization on all of the convolutional layers in YOLO we get more than 2\% improvement in mAP. Batch normalization also helps regularize the model. With batch normalization we can remove dropout from the model without overfitting.

\textbf{High Resolution Classifier.} All state-of-the-art detection methods use classifier pre-trained on ImageNet \cite{ILSVRC15}. Starting with AlexNet most classifiers operate on input images smaller than $256 \times 256$ \cite{alexnet}. The original YOLO trains the classifier network at $224 \times 224$ and increases the resolution to $448 \time 448$ for detection. This means the network has to simultaneously switch to learning object detection and adjust to the new input resolution.

For YOLOv2 we first fine tune the classification network at the full $448 \times 448$ resolution for 10 epochs on ImageNet. This gives the network time to adjust its filters to work better on higher resolution input. We then fine tune the resulting network on detection. This high resolution classification network gives us an increase of almost 4\% mAP.
   
\textbf{Convolutional With Anchor Boxes.} YOLO predicts the coordinates of bounding boxes directly using fully connected layers on top of the convolutional feature extractor. Instead of predicting coordinates directly Faster R-CNN predicts bounding boxes using hand-picked priors \cite{ren2015faster}. Using only convolutional layers the region proposal network (RPN) in Faster R-CNN predicts offsets and confidences for anchor boxes. Since the prediction layer is convolutional, the RPN predicts these offsets at every location in a feature map. Predicting offsets instead of coordinates simplifies the problem and makes it easier for the network to learn.

We remove the fully connected layers from YOLO and use anchor boxes to predict bounding boxes. First we eliminate one pooling layer to make the output of the network's convolutional layers higher resolution. We also shrink the network to operate on $416 \time 416$ input images instead of $448 \times 448$. We do this because we want an odd number of locations in our feature map so there is a single center cell. Objects, especially large objects, tend to occupy the center of the image so it's good to have a single location right at the center to predict these objects instead of four locations that are all nearby. YOLO's convolutional layers downsample the image by a factor of 32 so by using an input image of $416 \time 416$ we get an output feature map of $13 \times 13$.

When we move to anchor boxes we also decouple the class prediction mechanism from the spatial location and instead predict class and objectness for every anchor box. Following YOLO, the objectness prediction still predicts the IOU of the ground truth and the proposed box and the class predictions predict the conditional probability of that class given that there is an object.

Using anchor boxes we get a small decrease in accuracy. YOLO only predicts 98 boxes per image but with anchor boxes our model predicts more than a thousand. Without anchor boxes our intermediate model gets $69.5$ mAP with a recall of $81\%$. With anchor boxes our model gets $69.2$ mAP with a recall of $88\%$. Even though the mAP decreases, the increase in recall means that our model has more room to improve.

\textbf{Dimension Clusters.} We encounter two issues with anchor boxes when using them with YOLO. The first is that the box dimensions are hand picked. The network can learn to adjust the boxes appropriately but if we pick better priors for the network to start with we can make it easier for the network to learn to predict good detections.

Instead of choosing priors by hand, we run k-means clustering on the training set bounding boxes to automatically find good priors. If we use standard k-means with Euclidean distance larger boxes generate more error than smaller boxes. However, what we really want are priors that lead to good IOU scores, which is independent of the size of the box. Thus for our distance metric we use:

\[ d(\text{box}, \text{centroid}) = 1 - \text{IOU}(\text{box}, \text{centroid})\]

\begin{figure}[t]
      \centering
        \includegraphics[width=\linewidth]{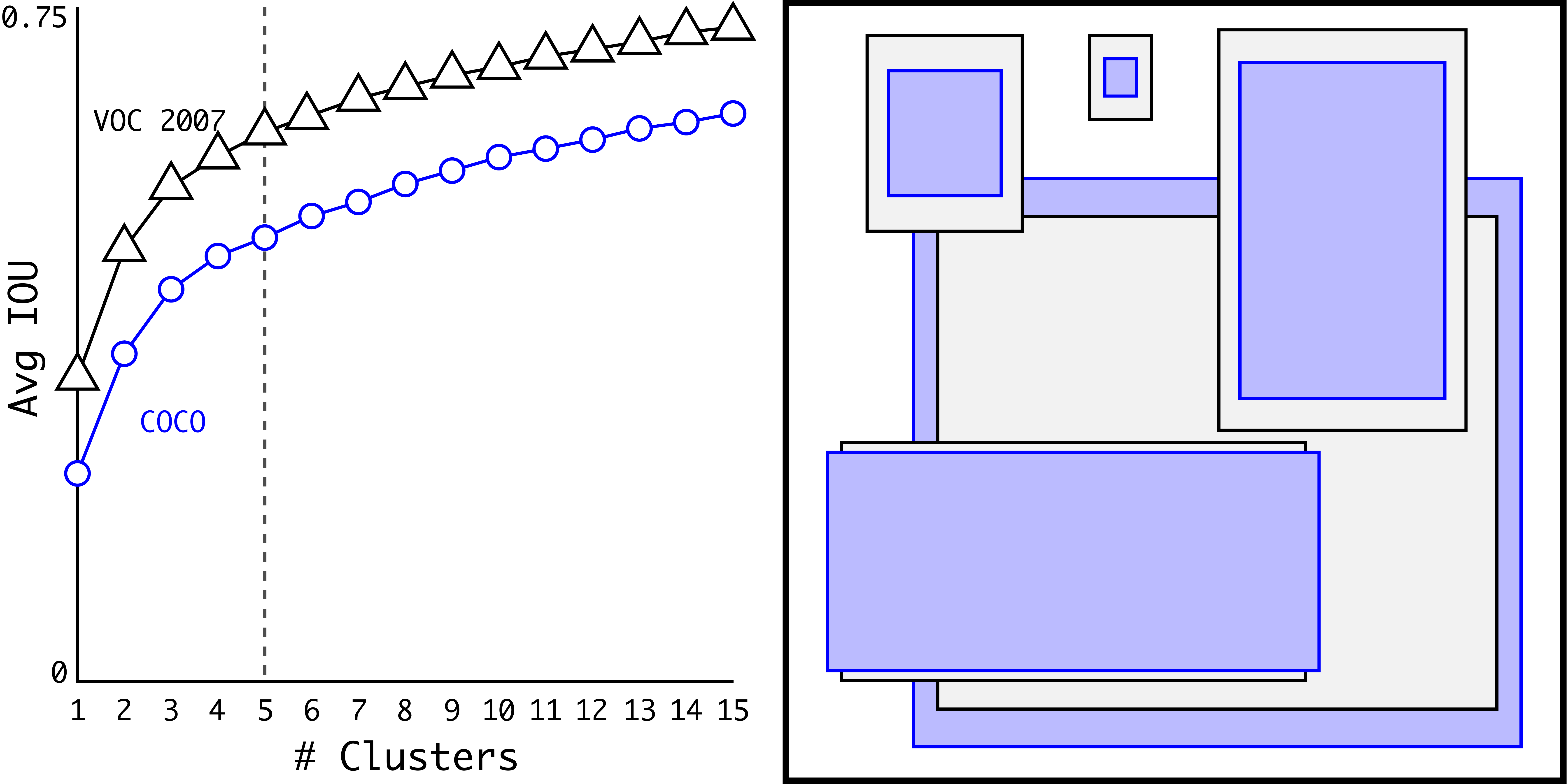}
      \caption{\small \textbf{Clustering box dimensions on VOC and COCO.} We run k-means clustering on the dimensions of bounding boxes to get good priors for our model. The left image shows the average IOU we get with various choices for $k$. We find that $k=5$ gives a good tradeoff for recall vs. complexity of the model. The right image shows the relative centroids for VOC and COCO. Both sets of priors favor thinner, taller boxes while COCO has greater variation in size than VOC. }
      \label{cluster}
   \end{figure}
   
We run k-means for various values of $k$ and plot the average IOU with closest centroid, see Figure \ref{cluster}. We choose $k=5$ as a good tradeoff between model complexity and high recall. The cluster centroids are significantly different than hand-picked anchor boxes. There are fewer short, wide boxes and more tall, thin boxes.

We compare the average IOU to closest prior of our clustering strategy and the hand-picked anchor boxes in Table \ref{boxes}. At only 5 priors the centroids perform similarly to 9 anchor boxes with an average IOU of 61.0 compared to 60.9. If we use 9 centroids we see a much higher average IOU. This indicates that using k-means to generate our bounding box starts the model off with a better representation and makes the task easier to learn.

\begin{table}[h]
\begin{center}
\begin{tabular}{lcc}
Box Generation & $\#$ & Avg IOU \\
\hline
Cluster SSE & 5 & 58.7 \\
Cluster IOU & 5 & 61.0 \\
Anchor Boxes \cite{ren2015faster}& 9 & 60.9 \\
Cluster IOU & 9 & 67.2 \\
\end{tabular}
\end{center}
\caption{\small \textbf{Average IOU of boxes to closest priors on VOC 2007.} The average IOU of objects on VOC 2007 to their closest, unmodified prior using different generation methods. Clustering gives much better results than using hand-picked priors.}
\label{boxes}
\end{table}

\textbf{Direct location prediction.} When using anchor boxes with YOLO we encounter a second issue: model instability, especially during early iterations. Most of the instability comes from predicting the $(x,y)$ locations for the box. In region proposal networks the network predicts values $t_x$ and $t_y$ and the $(x,y)$ center coordinates are calculated as:

\begin{align*}
x &= (t_x * w_a) - x_a \\
y &= (t_y * h_a) - y_a
\end{align*}

For example, a prediction of $t_x = 1$ would shift the box to the right by the width of the anchor box, a prediction of $t_x = -1$ would shift it to the left by the same amount.

This formulation is unconstrained so any anchor box can end up at any point in the image, regardless of what location predicted the box. With random initialization the model takes a long time to stabilize to predicting sensible offsets.

Instead of predicting offsets we follow the approach of YOLO and predict location coordinates relative to the location of the grid cell. This bounds the ground truth to fall between $0$ and $1$. We use a logistic activation to constrain the network's predictions to fall in this range.

The network predicts 5 bounding boxes at each cell in the output feature map. The network predicts 5 coordinates for each bounding box, $t_x$, $t_y$, $t_w$, $t_h$, and $t_o$. If the cell is offset from the top left corner of the image by $(c_x, c_y)$ and the bounding box prior has width and height $p_w$, $p_h$, then the predictions correspond to:

\begin{align*}
b_x &= \sigma(t_x) + c_x \\
b_y &= \sigma(t_y)  + c_y\\
b_w &= p_w e^{t_w}\\
b_h &= p_h e^{t_h}\\
Pr(\text{object}) * IOU(b, \text{object}) &= \sigma(t_o)\\
\end{align*}

Since we constrain the location prediction the parametrization is easier to learn, making the network more stable. Using dimension clusters along with directly predicting the bounding box center location improves YOLO by almost 5\% over the version with anchor boxes.

\begin{figure}[t]
      \centering
        \includegraphics[width=\linewidth]{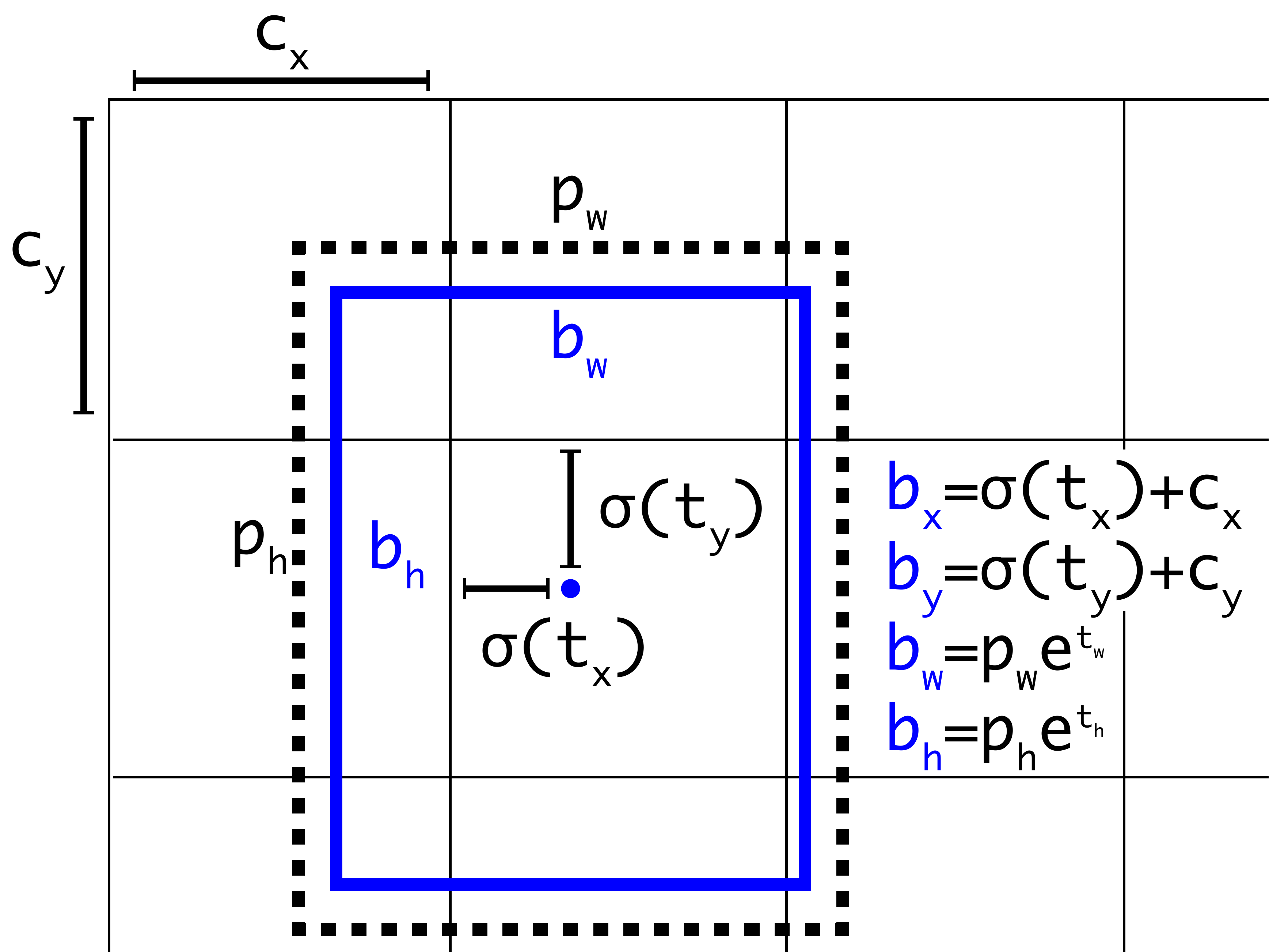}
      \caption{\small \textbf{Bounding boxes with dimension priors and location prediction.} We predict the width and height of the box as offsets from cluster centroids. We predict the center coordinates of the box relative to the location of filter application using a sigmoid function. }
      \label{box}
   \end{figure}

\textbf{Fine-Grained Features.}This modified YOLO predicts detections on a $13 \times 13$ feature map. While this is sufficient for large objects, it may benefit from finer grained features for localizing smaller objects. Faster R-CNN and SSD both run their proposal networks at various feature maps in the network to get a range of resolutions. We take a different approach, simply adding a passthrough layer that brings features from an earlier layer at $26 \times 26$ resolution.

The passthrough layer concatenates the higher resolution features with the low resolution features by stacking adjacent features into different channels instead of spatial locations, similar to the identity mappings in ResNet. This turns the $26 \times 26 \times 512$ feature map into a $13 \times 13 \times 2048$ feature map, which can be concatenated with the original features. Our detector runs on top of this expanded feature map so that it has access to fine grained features. This gives a modest 1\% performance increase.

\textbf{Multi-Scale Training.} The original YOLO uses an input resolution of $448 \times 448$. With the addition of anchor boxes we changed the resolution to $416 \times 416$. However, since our model only uses convolutional and pooling layers it can be resized on the fly. We want YOLOv2 to be robust to running on images of different sizes so we train this into the model.

Instead of fixing the input image size we change the network every few iterations. Every 10 batches our network randomly chooses a new image dimension size. Since our model downsamples by a factor of 32, we pull from the following multiples of 32: $\{320, 352, ... ,608\}$. Thus the smallest option is $320 \times 320$ and the largest is $608 \times 608$. We resize the network to that dimension and continue training. 

This regime forces the network to learn to predict well across a variety of input dimensions. This means the same network can predict detections at different resolutions. The network runs faster at smaller sizes so YOLOv2 offers an easy tradeoff between speed and accuracy.

At low resolutions YOLOv2 operates as a cheap, fairly accurate detector. At $288 \times 288$ it runs at more than 90 FPS with mAP almost as good as Fast R-CNN. This makes it ideal for smaller GPUs, high framerate video, or multiple video streams.

\begin{figure}[t]
      \centering
        \includegraphics[width=\linewidth]{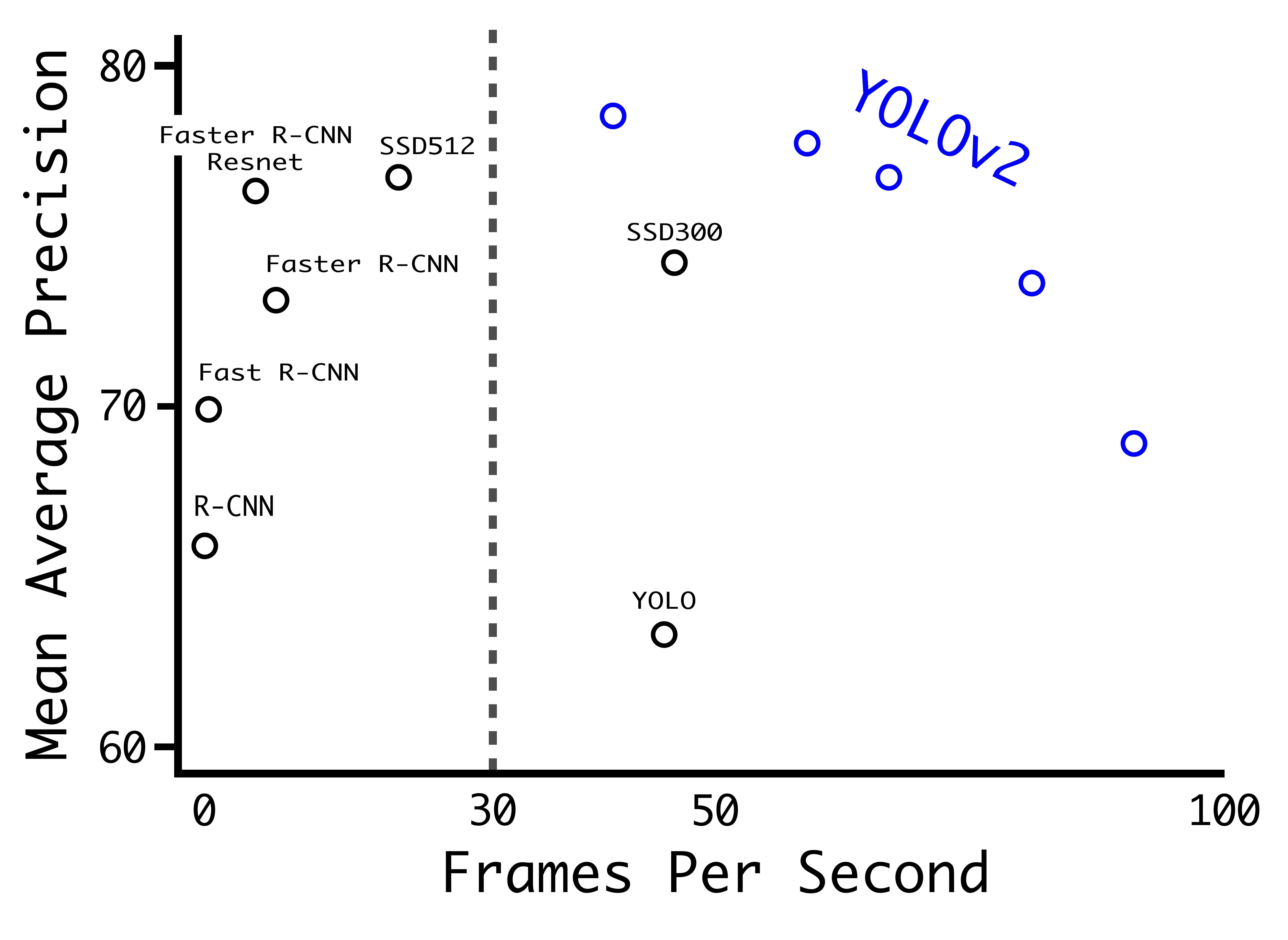}
      \caption{\small \textbf{Accuracy and speed on VOC 2007.}}
      \label{curve}
   \end{figure}

At high resolution YOLOv2 is a state-of-the-art detector with 78.6 mAP on VOC 2007 while still operating above real-time speeds. See Table \ref{timing} for a comparison of YOLOv2 with other frameworks on VOC 2007. Figure \ref{curve}

\textbf{Further Experiments.} We train YOLOv2 for detection on VOC 2012. Table \ref{voc12} shows the comparative performance of YOLOv2 versus other state-of-the-art detection systems. YOLOv2 achieves 73.4 mAP while running far faster than competing methods. We also train on COCO and compare to other methods in Table \ref{tab:coco}. On the VOC metric (IOU = .5) YOLOv2 gets 44.0 mAP, comparable to SSD and Faster R-CNN.

\begin{table*}
	\centering
	\setlength{\tabcolsep}{4pt}
	\begin{tabular}{r|c|ccccccc|c}
    	& YOLO & & & & & & & & YOLOv2 \\
    	\hline
        batch norm? & & \checkmark & \checkmark & \checkmark & \checkmark & \checkmark & \checkmark & \checkmark & \checkmark\\
        hi-res classifier? & & & \checkmark & \checkmark & \checkmark & \checkmark & \checkmark & \checkmark & \checkmark\\
        convolutional? & &  &  & \checkmark & \checkmark & \checkmark & \checkmark & \checkmark & \checkmark\\
        anchor boxes?  & &  &  & \checkmark & \checkmark &  &  & & \\
        new network?   & &  &  &  & \checkmark & \checkmark & \checkmark & \checkmark & \checkmark\\
        dimension priors?    & &  &  &  &  & \checkmark & \checkmark &  \checkmark &  \checkmark\\
        location prediction?    & &  &  &  &  & \checkmark & \checkmark &  \checkmark &  \checkmark\\

        passthrough?   & &  &  &  &  &  & \checkmark & \checkmark & \checkmark\\
        multi-scale?   & &  &  &  &  &  &  & \checkmark & \checkmark\\
        hi-res detector? & &  &  &  &  &  &  & & \checkmark\\
        \hline
        VOC2007 mAP & 63.4 & 65.8 & 69.5 & 69.2 & 69.6 & 74.4 & 75.4 & \rmed & \textbf{\rxl} \\
	\end{tabular}
    \caption{\textbf{The path from YOLO to YOLOv2.} Most of the listed design decisions lead to significant increases in mAP. Two exceptions are switching to a fully convolutional network with anchor boxes and using the new network. Switching to the anchor box style approach increased recall without changing mAP while using the new network cut computation by 33\%.}
    \label{analysis}
\end{table*}

\begin{table}[h]
\begin{center}
\begin{tabular}{lrrr}
Detection Frameworks & Train & mAP & FPS\\
\hline
Fast R-CNN \cite{fastrcnn}& 2007+2012 & 70.0 & 0.5 \\
Faster R-CNN VGG-16\cite{ren2015faster}& 2007+2012 & 73.2 & 7 \\
Faster R-CNN ResNet\cite{resnet}& 2007+2012 & 76.4 & 5 \\
YOLO \cite{yolo} & 2007+2012 & 63.4 & 45 \\
SSD300 \cite{ssd} & 2007+2012 & 74.3 & 46 \\
SSD500 \cite{ssd} & 2007+2012 & 76.8 & 19 \\
\hline
YOLOv2 $288 \times 288$ & 2007+2012 & \rxs & \sxs \\
YOLOv2 $352 \times 352$ & 2007+2012 & \rsmall & \ssmall \\
YOLOv2 $416 \times 416$ & 2007+2012 & \rmed & \smed \\
YOLOv2 $480 \times 480$ & 2007+2012 & \rlarge & \slarge \\
YOLOv2 $544 \times 544$ & 2007+2012 & \textbf{\rxl} & \sxl \\
\end{tabular}
\end{center}
\caption{\small \textbf{Detection frameworks on \textsc{Pascal} VOC 2007.} YOLOv2 is faster and more accurate than prior detection methods. It can also run at different resolutions for an easy tradeoff between speed and accuracy. Each YOLOv2 entry is actually the same trained model with the same weights, just evaluated at a different size. All timing information is on a Geforce GTX Titan X (original, not Pascal model).}
\label{timing}
\end{table}

\begin{table*}[ht]\scriptsize
	\centering
	\setlength{\tabcolsep}{1.45pt}
	\begin{tabular}{l|c|c|cccccccccccccccccccc}
		Method & data &  mAP &  aero &  bike &  bird &  boat &  bottle &  bus &  car &  cat &  chair &  cow &  table &  dog &  horse &  mbike &  person &  plant &  sheep &  sofa &  train &  tv \\
        \hline
		Fast R-CNN \cite{fastrcnn} & 07++12 & 68.4 & 82.3 & 78.4 & 70.8 & 52.3 & 38.7 & 77.8 & 71.6 & 89.3 & 44.2 & 73.0 & 55.0 & 87.5 & 80.5 & 80.8 & 72.0 & 35.1 & 68.3 & 65.7 & 80.4 & 64.2\\
		Faster R-CNN \cite{ren2015faster} & 07++12 & 70.4 & 84.9 & 79.8 & 74.3 & 53.9 & 49.8 & 77.5 & 75.9 & 88.5 & 45.6 & 77.1 & 55.3 & 86.9 & 81.7 & 80.9 & 79.6 & 40.1 & 72.6 & 60.9 & 81.2 & 61.5\\
		YOLO \cite{yolo} & 07++12 & 57.9 & 77.0 & 67.2 & 57.7 & 38.3 & 22.7 & 68.3 & 55.9 & 81.4 & 36.2 & 60.8 & 48.5 & 77.2 & 72.3 & 71.3 & 63.5 & 28.9 & 52.2 & 54.8 & 73.9 & 50.8\\
        SSD300 \cite{ssd} & 07++12 & 72.4 & 85.6 & 80.1 & 70.5 & 57.6 & 46.2 & 79.4 & 76.1 & 89.2 & 53.0 & 77.0 & 60.8 & 87.0 & 83.1 & 82.3 & 79.4 & 45.9 & 75.9 & 69.5 & 81.9 & 67.5\\
        SSD512 \cite{ssd} & 07++12 & 74.9 & 87.4 & 82.3 & 75.8 & 59.0 & 52.6 & 81.7 & 81.5 & 90.0 & 55.4 & 79.0 & 59.8 & 88.4 & 84.3 & 84.7 & 83.3 & 50.2 & 78.0 & 66.3 & 86.3 & 72.0\\
        ResNet \cite{resnet} & 07++12 & 73.8 & 86.5 &81.6 &77.2 &58.0 &51.0 &78.6 &76.6 &93.2 &48.6 &80.4 &59.0 &92.1 &85.3 &84.8 &80.7 &48.1 &77.3 &66.5 &84.7 &65.6\\
        \hline
        YOLOv2 $544$& 07++12 & 73.4 & 86.3 & 82.0 & 74.8 & 59.2 & 51.8 & 79.8 & 76.5 & 90.6 & 52.1 & 78.2 & 58.5 & 89.3 & 82.5 & 83.4 & 81.3 & 49.1 & 77.2 & 62.4 & 83.8 & 68.7\\

        \noalign{\smallskip}
	\end{tabular}
	\caption{\textbf{PASCAL VOC2012 \texttt{test} detection results.} YOLOv2 performs on par with state-of-the-art detectors like Faster R-CNN with ResNet and SSD512 and is $2-10\times$ faster.}
    \label{voc12}
\end{table*}

\begin{table*}
	\centering
	\setlength{\tabcolsep}{1pt}
	\begin{tabular}{l|c|ccc|ccc|ccc|ccc}
        & & 0.5:0.95 & 0.5 & 0.75 & S & M & L & 1 & 10 & 100 & S & M & L\\
        \hline
        Fast R-CNN \cite{fastrcnn} & train & 19.7 & 35.9 & - & - & - & - & - & - & - & - & - & -\\
        Fast R-CNN\cite{bell2015inside} & train & 20.5 & 39.9 & 19.4 & 4.1 & 20.0 & 35.8 & 21.3 & 29.5 & 30.1 & 7.3 & 32.1 & 52.0\\
        Faster R-CNN\cite{ren2015faster} & trainval & 21.9 & 42.7 & - & - & - & - & - & - & - & - & - & -\\
        ION \cite{bell2015inside} & train & 23.6 & 43.2 & 23.6 & 6.4 & 24.1 & 38.3 & 23.2 & 32.7 & 33.5 & 10.1 & 37.7 & 53.6\\
        Faster R-CNN\cite{coco} & trainval & 24.2 & 45.3 & 23.5 & 7.7 & 26.4 & 37.1 & 23.8 & 34.0 & 34.6 & 12.0 & 38.5 & 54.4\\
        SSD300 \cite{ssd} & trainval35k & 23.2 & 41.2 & 23.4 & 5.3 & 23.2 & 39.6 & 22.5 & 33.2 & 35.3 & 9.6 & 37.6 & 56.5\\
        SSD512 \cite{ssd}& trainval35k & \textbf{26.8} & \textbf{46.5} & \textbf{27.8} & \textbf{9.0} & \textbf{28.9} & \textbf{41.9} & \textbf{24.8} & \textbf{37.5} & \textbf{39.8} & \textbf{14.0} & \textbf{43.5} & \textbf{59.0}\\
        \hline
        YOLOv2 \cite{ssd} & trainval35k & 21.6 & 44.0 & 19.2 & 5.0 & 22.4 & 35.5 & 20.7 & 31.6 & 33.3 & 9.8 & 36.5 & 54.4\\
        
    \end{tabular}
    \caption{\textbf{Results on COCO \texttt{test-dev2015}. Table adapted from \cite{ssd}}}
    \label{tab:coco}
\end{table*}

\section{Faster}

We want detection to be accurate but we also want it to be fast. Most applications for detection, like robotics or self-driving cars, rely on low latency predictions. In order to maximize performance we design YOLOv2 to be fast from the ground up.

Most detection frameworks rely on VGG-16 as the base feature extractor \cite{vgg}. VGG-16 is a powerful, accurate classification network but it is needlessly complex. The convolutional layers of VGG-16 require 30.69 billion floating point operations for a single pass over a single image at $224 \times 224$ resolution.

The YOLO framework uses a custom network based on the Googlenet architecture \cite{googlenet}. This network is faster than VGG-16, only using 8.52 billion operations for a forward pass. However, it's accuracy is slightly worse than VGG-16. For single-crop, top-5 accuracy at $224 \times 224$, YOLO's custom model gets 88.0\% ImageNet compared to 90.0\% for VGG-16.

\textbf{Darknet-19}. We propose a new classification model to be used as the base of YOLOv2. Our model builds off of prior work on network design as well as common knowledge in the field. Similar to the VGG models we use mostly $3 \times 3$ filters and double the number of channels after every pooling step \cite{vgg}. Following the work on Network in Network (NIN) we use global average pooling to make predictions as well as $1 \times 1$ filters to compress the feature representation between $3 \times 3$ convolutions \cite{lin2013network}. We use batch normalization to stabilize training, speed up convergence, and regularize the model \cite{batch}.

Our final model, called Darknet-19, has 19 convolutional layers and 5 maxpooling layers. For a full description see Table \ref{net}. Darknet-19 only requires 5.58 billion operations to process an image yet achieves $72.9\%$ top-1 accuracy and $91.2\%$ top-5 accuracy on ImageNet. 

\textbf{Training for classification.} We train the network on the standard ImageNet 1000 class classification dataset for 160 epochs using stochastic gradient descent with a starting learning rate of $0.1$, polynomial rate decay with a power of $4$, weight decay of $0.0005$ and momentum of $0.9$ using the Darknet neural network framework \cite{darknet13}. During training we use standard data augmentation tricks including random crops, rotations, and hue, saturation, and exposure shifts.

\begin{table}[h] \scriptsize
\begin{center}
\begin{tabular}{c|c|c|c}
Type & Filters & Size/Stride & Output\\
\hline
Convolutional & 32 & $3 \times 3$ & $224 \times 224 $ \\
Maxpool & &$2 \times 2 / 2$ & $112 \times 112 $ \\
Convolutional & 64 & $3 \times 3$ & $112 \times 112 $ \\
Maxpool & & $2 \times 2 / 2$ & $56 \times 56 $ \\
Convolutional & 128 &$3 \times 3$ & $56 \times 56 $ \\
Convolutional & 64 &$1 \times 1$ & $56 \times 56 $ \\
Convolutional & 128 &$3 \times 3$ & $56 \times 56 $ \\
Maxpool & & $2 \times 2 / 2$ & $28 \times 28 $ \\
Convolutional & 256 & $3 \times 3$ & $28 \times 28 $ \\
Convolutional & 128 & $1 \times 1$ & $28 \times 28 $ \\
Convolutional & 256& $3 \times 3$ & $28 \times 28 $ \\
Maxpool & & $2 \times 2 / 2$ & $14 \times 14 $ \\
Convolutional & 512 & $3 \times 3$ & $14 \times 14 $ \\
Convolutional & 256& $1 \times 1$ & $14 \times 14 $ \\
Convolutional & 512 & $3 \times 3$ & $14 \times 14$ \\
Convolutional & 256& $1 \times 1$ & $14 \times 14$ \\
Convolutional & 512 & $3 \times 3$ & $14 \times 14 $ \\
Maxpool & & $2 \times 2 / 2$ & $7 \times 7 $ \\
Convolutional & 1024 & $3 \times 3$ & $7 \times 7 $ \\
Convolutional & 512 & $1 \times 1$ & $7 \times 7 $ \\
Convolutional & 1024 & $3 \times 3$ & $7 \times 7$ \\
Convolutional & 512 & $1 \times 1$ & $7 \times 7$ \\
Convolutional & 1024 & $3 \times 3$ & $7 \times 7$ \\
\hline
\hline
Convolutional & 1000 & $1 \times 1$ & $7 \times 7$ \\
Avgpool & & Global & $1000$ \\
Softmax & & &\\
\end{tabular}
\end{center}
\caption{\small \textbf{Darknet-19.}}
\label{net}
\end{table}

As discussed above, after our initial training on images at $224 \times 224$ we fine tune our network at a larger size, $448 \time 448$. For this fine tuning we train with the above parameters but for only 10 epochs and starting at a learning rate of $10^{-3}$. At this higher resolution our network achieves a top-1 accuracy of $76.5\%$ and a top-5 accuracy of $93.3\%$.

\textbf{Training for detection.} We modify this network for detection by removing the last convolutional layer and instead adding on three $3 \times 3$ convolutional layers with $1024$ filters each followed by a final $1\times 1$ convolutional layer with the number of outputs we need for detection. For VOC we predict 5 boxes with 5 coordinates each and 20 classes per box so 125 filters. We also add a passthrough layer from the final $3 \times 3 \times 512$ layer to the second to last convolutional layer so that our model can use fine grain features.

We train the network for 160 epochs with a starting learning rate of $10^{-3}$, dividing it by 10 at 60 and 90 epochs. We use a weight decay of $0.0005$ and momentum of $0.9$. We use a similar data augmentation to YOLO and SSD with random crops, color shifting, etc. We use the same training strategy on COCO and VOC.

\section{Stronger}

We propose a mechanism for jointly training on classification and detection data. Our method uses images labelled for detection to learn detection-specific information like bounding box coordinate prediction and objectness as well as how to classify common objects. It uses images with only class labels to expand the number of categories it can detect.

During training we mix images from both detection and classification datasets. When our network sees an image labelled for detection we can backpropagate based on the full YOLOv2 loss function. When it sees a classification image we only backpropagate loss from the classification-specific parts of the architecture.

This approach presents a few challenges. Detection datasets have only common objects and general labels, like ``dog'' or ``boat''.  Classification datasets have a much wider and deeper range of labels. ImageNet has more than a hundred breeds of dog, including ``Norfolk terrier'', ``Yorkshire terrier'', and ``Bedlington terrier''. If we want to train on both datasets we need a coherent way to merge these labels.

Most approaches to classification use a softmax layer across all the possible categories to compute the final probability distribution. Using a softmax assumes the classes are mutually exclusive. This presents problems for combining datasets, for example you would not want to combine ImageNet and COCO using this model because the classes ``Norfolk terrier'' and ``dog'' are not mutually exclusive.

We could instead use a multi-label model to combine the datasets which does not assume mutual exclusion. This approach ignores all the structure we do know about the data, for example that all of the COCO classes are mutually exclusive.

\textbf{Hierarchical classification.} ImageNet labels are pulled from WordNet, a language database that structures concepts and how they relate \cite{wordnet}. In WordNet, ``Norfolk terrier'' and ``Yorkshire terrier'' are both hyponyms of ``terrier'' which is a type of ``hunting dog'', which is a type of ``dog'', which is a ``canine'', etc. Most approaches to classification assume a flat structure to the labels however for combining datasets, structure is exactly what we need.

WordNet is structured as a directed graph, not a tree, because language is complex. For example a ``dog'' is both a type of ``canine'' and a type of ``domestic animal'' which are both synsets in WordNet. Instead of using the full graph structure, we simplify the problem by building a hierarchical tree from the concepts in ImageNet.

To build this tree we examine the visual nouns in ImageNet and look at their paths through the WordNet graph to the root node, in this case ``physical object''. Many synsets only have one path through the graph so first we add all of those paths to our tree. Then we iteratively examine the concepts we have left and add the paths that grow the tree by as little as possible. So if a concept has two paths to the root and one path would add three edges to our tree and the other would only add one edge, we choose the shorter path.

The final result is WordTree, a hierarchical model of visual concepts. To perform classification with WordTree we predict conditional probabilities at every node for the probability of each hyponym of that synset given that synset. For example, at the ``terrier'' node we predict:

\begin{align*}
Pr(\text{Norfolk terrier} &| \text{terrier}) \\
Pr(\text{Yorkshire terrier} &| \text{terrier}) \\
Pr(\text{Bedlington terrier} &| \text{terrier})\\
...&\\
\end{align*}

If we want to compute the absolute probability for a particular node we simply follow the path through the tree to the root node and multiply to conditional probabilities. So if we want to know if a picture is of a Norfolk terrier we compute:

\begin{align*}
Pr(\text{Norfolk terrier}) &= Pr(\text{Norfolk terrier} | \text{terrier})\\
 * Pr(\text{terrier} &| \text{hunting dog}) \\
* \ldots& * \\
*Pr(\text{mammal} &| Pr(\text{animal})\\
 * Pr(\text{animal} &| \text{physical object})
\end{align*}

For classification purposes we assume that the the image contains an object: $Pr(\text{physical object}) = 1$. 

\begin{figure}[t]
      \centering
        \includegraphics[width=\linewidth]{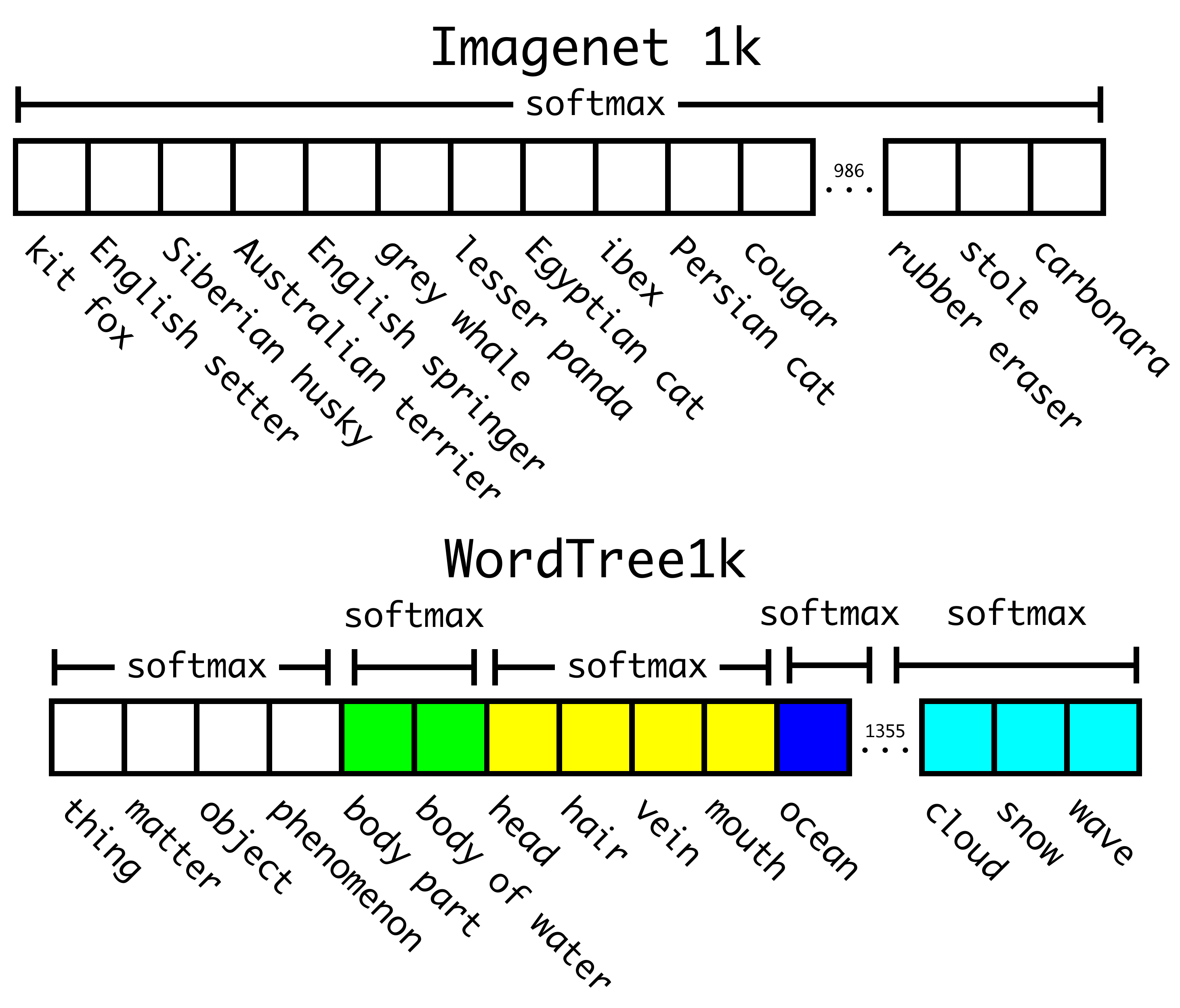}
      \caption{\small \textbf{Prediction on ImageNet vs WordTree.} Most ImageNet models use one large softmax to predict a probability distribution. Using WordTree we perform multiple softmax operations over co-hyponyms.}
      \label{softmax}
   \end{figure}

To validate this approach we train the Darknet-19 model on WordTree built using the 1000 class ImageNet. To build WordTree1k we add in all of the intermediate nodes which expands the label space from 1000 to 1369. During training we propagate ground truth labels up the tree so that if an image is labelled as a ``Norfolk terrier'' it also gets labelled as a ``dog'' and a ``mammal'', etc. To compute the conditional probabilities our model predicts a vector of 1369 values and we compute the softmax over all sysnsets that are hyponyms of the same concept, see Figure \ref{softmax}.

Using the same training parameters as before, our hierarchical Darknet-19 achieves $71.9\%$ top-1 accuracy and $90.4\%$ top-5 accuracy. Despite adding 369 additional concepts and having our network predict a tree structure our accuracy only drops marginally. Performing classification in this manner also has some benefits. Performance degrades gracefully on new or unknown object categories. For example, if the network sees a picture of a dog but is uncertain what type of dog it is, it will still predict ``dog'' with high confidence but have lower confidences spread out among the hyponyms.

This formulation also works for detection. Now, instead of assuming every image has an object, we use YOLOv2's objectness predictor to give us the value of $Pr(\text{physical object})$. The detector predicts a bounding box and the tree of probabilities. We traverse the tree down, taking the highest confidence path at every split until we reach some threshold and we predict that object class.

\begin{figure}[t]
      \centering
        \includegraphics[width=\linewidth]{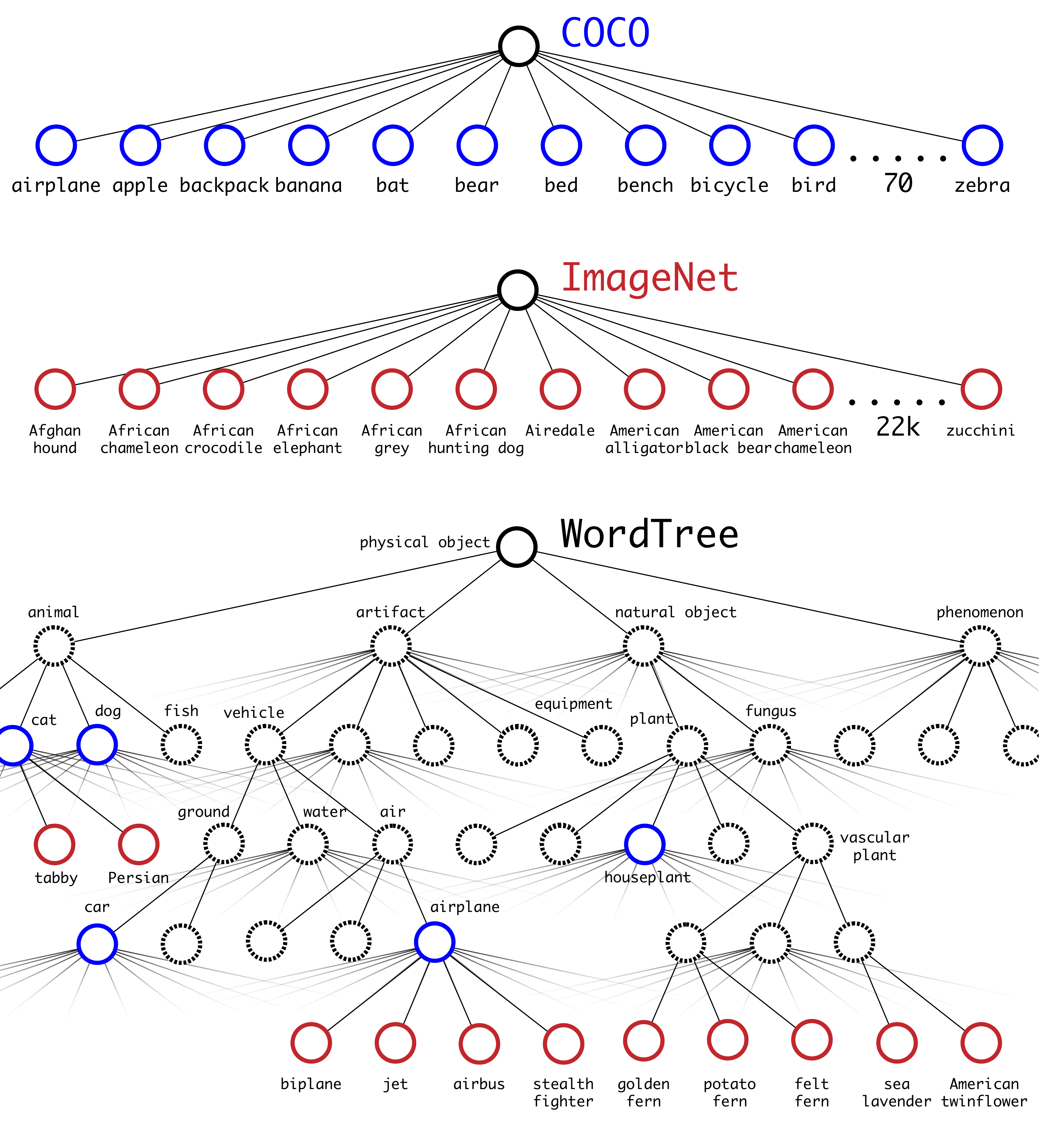}
      \caption{\small \textbf{Combining datasets using WordTree hierarchy.} Using the WordNet concept graph we build a hierarchical tree of visual concepts. Then we can merge datasets together by mapping the classes in the dataset to synsets in the tree. This is a simplified view of WordTree for illustration purposes.}
      \label{tree}
   \end{figure}

\textbf{Dataset combination with WordTree.} We can use WordTree to combine multiple datasets together in a sensible fashion. We simply map the categories in the datasets to synsets in the tree. Figure \ref{tree} shows an example of using WordTree to combine the labels from ImageNet and COCO. WordNet is extremely diverse so we can use this technique with most datasets.

\textbf{Joint classification and detection.} Now that we can combine datasets using WordTree we can train our joint model on classification and detection. We want to train an extremely large scale detector so we create our combined dataset using the COCO detection dataset and the top 9000 classes from the full ImageNet release. We also need to evaluate our method so we add in any classes from the ImageNet detection challenge that were not already included. The corresponding WordTree for this dataset has 9418 classes. ImageNet is a much larger dataset so we balance the dataset by oversampling COCO so that ImageNet is only larger by a factor of 4:1.

Using this dataset we train YOLO9000. We use the base YOLOv2 architecture but only 3 priors instead of 5 to limit the output size. When our network sees a detection image we backpropagate loss as normal. For classification loss, we only backpropagate loss at or above the corresponding level of the label. For example, if the label is ``dog'' we do assign any error to predictions further down in the tree, ``German Shepherd'' versus ``Golden Retriever'', because we do not have that information.

When it sees a classification image we only backpropagate classification loss. To do this we simply find the bounding box that predicts the highest probability for that class and we compute the loss on just its predicted tree. We also assume that the predicted box overlaps what would be the ground truth label by at least $.3$ IOU and we backpropagate objectness loss based on this assumption.

Using this joint training, YOLO9000 learns to find objects in images using the detection data in COCO and it learns to classify a wide variety of these objects using data from ImageNet.

We evaluate YOLO9000 on the ImageNet detection task. The detection task for ImageNet shares on 44 object categories with COCO which means that YOLO9000 has only seen classification data for the majority of the test images, not detection data. YOLO9000 gets \inet{} mAP overall with \inone{} mAP on the disjoint 156 object classes that it has never seen any labelled detection data for. This mAP is higher than results achieved by DPM but YOLO9000 is trained on different datasets with only partial supervision \cite{DPM}. It also is simultaneously detecting 9000 other object categories, all in real-time.

\begin{table}[h]
\begin{center}
\begin{tabular}{lc}
diaper & 0.0\\
horizontal bar & 0.0 \\
rubber eraser & 0.0 \\
sunglasses & 0.0 \\
swimming trunks & 0.0 \\
... \\
red panda & 50.7 \\
fox  & 52.1  \\
koala bear & 54.3   \\
tiger  & 61.0  \\
armadillo & 61.7\\

\end{tabular}
\end{center}
\caption{\small \textbf{YOLO9000 Best and Worst Classes on ImageNet.} The classes with the highest and lowest AP from the 156 weakly supervised classes. YOLO9000 learns good models for a variety of animals but struggles with new classes like clothing or equipment.}
\label{res}
\end{table}

When we analyze YOLO9000's performance on ImageNet we see it learns new species of animals well but struggles with learning categories like clothing and equipment. New animals are easier to learn because the objectness predictions generalize well from the animals in COCO. Conversely, COCO does not have bounding box label for any type of clothing, only for person, so YOLO9000 struggles to model categories like ``sunglasses'' or ``swimming trunks''.

\section{Conclusion}

We introduce YOLOv2 and YOLO9000, real-time detection systems. YOLOv2 is state-of-the-art and faster than other detection systems across a variety of detection datasets. Furthermore, it can be run at a variety of image sizes to provide a smooth tradeoff between speed and accuracy.

YOLO9000 is a real-time framework for detection more than 9000 object categories by jointly optimizing detection and classification. We use WordTree to combine data from various sources and our joint optimization technique to train simultaneously on ImageNet and COCO. YOLO9000 is a strong step towards closing the dataset size gap between detection and classification.

Many of our techniques generalize outside of object detection. Our WordTree representation of ImageNet offers a richer, more detailed output space for image classification. Dataset combination using hierarchical classification would be useful in the classification and segmentation domains. Training techniques like multi-scale training could provide benefit across a variety of visual tasks.

 For future work we hope to use similar techniques for weakly supervised image segmentation. We also plan to improve our detection results using more powerful matching strategies for assigning weak labels to classification data during training. Computer vision is blessed with an enormous amount of labelled data. We will continue looking for ways to bring different sources and structures of data together to make stronger models of the visual world.

{\small
\bibliographystyle{ieee}
\bibliography{yolo}
}

\end{document}